\documentclass{article}

\usepackage[preprint]{neurips_2023}

\usepackage[utf8]{inputenc} 
\usepackage[T1]{fontenc}    
\usepackage{hyperref}       
\usepackage{url}            
\usepackage{booktabs}       
\usepackage{amsfonts}       
\usepackage{nicefrac}       
\usepackage{microtype}      
\usepackage{xcolor}         
\usepackage{amsmath}
\usepackage{natbib}

\title{Reverse Engineering Deep ReLU Networks An Optimization-based Algorithm}

\author{%
  Mehrab Hamidi \\
  McGill University\\
  Mila Quebec AI Institute \\
  \texttt{mehrab.hamidi@mila.quebec} \\
}

\begin{document}

\maketitle

\begin{abstract}
Reverse engineering deep ReLU networks is a critical problem in understanding the complex behavior and interpretability of neural networks. In this research, we present a novel method for reconstructing deep ReLU networks by leveraging convex optimization techniques and a sampling-based approach. Our method begins by sampling points in the input space and querying the black box model to obtain the corresponding hyperplanes. We then define a convex optimization problem with carefully chosen constraints and conditions to guarantee its convexity. The objective function is designed to minimize the discrepancy between the reconstructed network's output and the target model's output, subject to the constraints. We employ gradient descent to optimize the objective function, incorporating L1 or L2 regularization as needed to encourage sparse or smooth solutions. Our research contributes to the growing body of work on reverse engineering deep ReLU networks and paves the way for new advancements in neural network interpretability and security.
\end{abstract}

\section{Introduction and Motivation}

Deep learning has established itself as a powerful tool across various domains, including computer vision, natural language processing, and reinforcement learning. Among different types of deep neural networks, Rectified Linear Unit (ReLU) networks have gained particular popularity due to their simplicity and strong expressive capabilities. These networks employ piece-wise linear activation functions, allowing them to efficiently approximate complex functions and provide a robust learning mechanism. Despite the widespread success of deep ReLU networks, understanding their properties, expressive power, and the feasibility of reverse-engineering their structure remains a significant challenge in the field. \\
Reverse engineering deep ReLU networks can serve multiple purposes. First, it can facilitate a better understanding of the inner workings of these models, enabling researchers to devise improved architectures, training methods, and optimization techniques. Second, it can promote interpretability and robustness, helping to identify and mitigate potential vulnerabilities and security risks associated with deep learning models. However, reverse engineering deep ReLU networks presents significant challenges, as it involves determining the optimal weights and biases for the networks given limited information. Hence, Reverse-engineering deep ReLU networks is a critical task, as it can offer insights into the model's inner workings, contribute to the development of more efficient and accurate models, and uncover potential vulnerabilities associated with the model's architecture and weights. In this context, numerous research efforts have sought to investigate the properties of deep ReLU networks, their expressive power, and their reverse-engineering potential. \\
In this work, we propose a novel approach to reverse engineering deep ReLU networks by formulating an optimization problem and employing gradient descent to solve it. We build upon insights from previous works on the complexity of linear regions and activation patterns in deep ReLU networks, the expressive power of deep neural networks, and methods for model reconstruction from model explanations. Our approach aims to advance the understanding of deep ReLU networks and provide a practical method for reconstructing them from limited information.


\section{Related Work}
Our work builds upon a rich body of research in the areas of deep learning, neural networks, and their properties. 
\cite{fefferman1994reconstructing} is probably the first one, who explored the problem of reconstructing a neural net from its output, showing that under certain conditions, it is possible to recover the network's structure and parameters from its input-output mapping. This work provided an early theoretical foundation for the study of reverse-engineering neural networks and highlighted the challenges and limitations of this task, such as the non-uniqueness of solutions and the need for regularization techniques to stabilize the reconstruction process.
\cite{rolnick2020reverse} focused on reverse-engineering deep ReLU networks by leveraging the piecewise-linear property of ReLU activations. Their approach required querying the input-output mapping of the model, and their algorithm recovered the network layer by layer. Their method assumed that the weights are integer valued and the network is noiseless.
\cite{carlini2020cryptanalytic} investigated the extraction of neural network models in a cryptanalytic setting, demonstrating the feasibility of recovering model parameters by querying the model's outputs. Their method used a black-box query oracle and was based on solving a system of linear equations. The approach made minimal assumptions about the network architecture and activation functions. However, it required a substantial number of queries, and the exact architecture of the network was not recovered.
Both \cite{carlini2020cryptanalytic} and \cite{rolnick2020reverse} propose algorithms for reverse-engineering deep ReLU networks, with \cite{carlini2020cryptanalytic} assuming known architecture and \cite{rolnick2020reverse} being more efficient in terms of time complexity despite needing much more queries.
\cite{milli2019model} explored the related problem of model reconstruction from model explanations, showing that it is possible to recover a model from explanations such as feature importance scores or model predictions. They leveraged model explanations like LIME and SHAP to recover the target model's structure and parameters. The method's accuracy depended on the quality of the explanations and the choice of the explanation method. They did not focus specifically on ReLU networks, and the method was applicable to various types of models. However, Their work is connected to our problem of reverse engineering deep ReLU networks, as both problems involve reconstructing model parameters from limited information.
\cite{simon2022reverse} investigated the reverse engineering of the Neural Tangent Kernel (NTK), a widely used tool in the analysis of deep learning dynamics. They showed that the NTK can be recovered from a model's input-output pairs, providing a novel approach to understanding the inner workings of deep networks. This work contributes to the growing body of research on reverse-engineering deep learning models and offers a new perspective on the relationship between the NTK and the model's underlying structure.
The algorithms for reconstruction of networks have limitations, for example, \cite{rolnick2020reverse} in terms of scalability and handling complex architectures or \cite{carlini2020cryptanalytic}'s  algorithm assumes that the architecture of the network is known, which is not always the case in practical applications but it requires less queries from the input-output and as a result it has a more efficient algorithm in sense of space-complexity whilst \cite{rolnick2020reverse} is a more efficient algorithm in the sense of time complexity.\\
In contrast to the above methods, other works in the related literature focused on understanding the expressive power and properties of deep ReLU networks.  Despite recent works \cite{zhang2021understanding} \cite{lin2017does} \cite{hanin2019complexity} \cite{haninrolnick:2019} investigate the expressive power of deep neural networks, there are lots of unknown and open problems on that regards.
\cite{haninrolnick:2019} performed an analysis of the activation patterns of deep ReLU networks, showing that these networks exhibit surprisingly few activation patterns relative to the number of neurons. Their findings offered insights into the expressive power of deep ReLU networks and the complexity of their input-output mappings, revealing an inherent simplicity that may contribute to the effectiveness of these models. They also provided bounds on the number of activation patterns, which can be useful for understanding the behavior of deep ReLU networks, similarly \cite{hanin2019complexity} investigated the complexity of linear regions in deep networks, shedding light on the geometry of the decision boundaries formed by ReLU networks. Their work provided a rigorous understanding of the underlying mathematical structure of these models and the factors that contribute to their expressive power. They showed that the number of linear regions grows exponentially with the depth of the network, highlighting the role of depth in enabling complex decision boundaries. These insights of \cite{haninrolnick:2019} \cite{hanin2019complexity} inform our understanding of the expressive power of deep ReLU networks and motivate our approach to reverse engineering them. 
In another work \cite{raghu2017expressive} investigated the expressive power of deep neural networks, providing insights into the capacity of deep networks to learn complex functions and demonstrating that certain network architectures are better suited for learning specific types of functions. Their work provided insights into the role of depth, width, and architecture in determining the expressiveness of deep learning models. They developed a framework for understanding the expressiveness of networks in terms of their ability to represent functions and transformations. Similarly, \cite{hanin2019universal} prove that for any continuous function on a compact set, there exists a deep neural network with a bounded width and ReLU activation functions that can approximate the function to within any desired degree of accuracy and highlights potential of these kind of networks for efficient and accurate representation of complex functions. Hence these works helps us understand the inherent expressive power of deep ReLU networks and guides our choice of network architecture for the reverse engineering problem. One other similar work of \cite{rolnick2017power} examined the power of deeper networks for expressing natural functions, showing that deeper networks can learn a broader class of functions compared to shallow networks. This work further supports the idea that deep ReLU networks are capable of expressing complex functions and serves as a motivation for our study on reverse engineering such networks.
\cite{bui2020functional} examined the functional versus parametric equivalence of ReLU networks, exploring the relationship between the architecture of a network and its ability to represent functions. They showed that functionally equivalent networks can have vastly different architectures and parameters, underscoring the importance of studying functional properties rather than specific parameter settings. Their work also provided a framework for understanding the space of functionally equivalent networks, which can inform the design of more efficient and robust models. \\
In summary, the related work presented here highlights the rich landscape of research efforts aimed at understanding and reverse-engineering deep ReLU networks. These studies have provided valuable insights into the properties, expressive power, and vulnerabilities of these models, paving the way for further advancements in the field. By building upon these foundational works, including Fefferman's early exploration of the problem in 1994, we hope to inspire new research directions and contribute to the ongoing quest to unravel the mysteries of deep learning.

\section{Background}

    \textbf{Def.} The rectified linear unit function ReLU:$\mathbb{R} \rightarrow \mathbb{R}$ is defined by $ReLU(x) = max\{0, x\}$. For any $n \in \mathbb{N}$ we denote the non-linearity $\sigma$, the map $\sigma:\mathbb{R}^n \rightarrow \mathbb{R}^n$, that applier ReLU to each coordinate. \\
    \textbf{Def.} A real-valued feed-forward Deep ReLU network with architecture $n_0, n_1, \cdots, n_m$ ($n_m = 1$) is a ordered collection of affine maps $A^1, A^2, ..., A^m$ such that 
    \[
    A^{i} : \mathbb{R}^{n_{i - 1}} \rightarrow \mathbb{R}^{n_{i}} ; \forall i = 1, \cdots, m \\
    \]
    Or equivalently we determine $A^{i}$ affine map by a weight matrix $W^i$ with dimension $n_i \times n_{i-1}$ and bias vector $b^i$, such that
    \[
    [W^i | b^i][x, 1]^T = A^i(x)
    \]
    We show this deep ReLU network by function $f(x)$ as follows:
    \[
    f(x) := \sigma \circ A^m \circ \sigma \circ A^{m-1} \circ \cdots \circ \sigma \circ A^1(x)
    \]
    \textbf{Def.} The parameter space of this neural network is denoted by 
    \[
    \begin{array}{l}
        \mathcal{P}(n_0, \cdots, n_m):= \mathbb{R}^{D(n_0, \cdots,n_m)} \\
        D(n_0, \cdots,n_m) := \sum_{i=1}^m n_i (n_{i-1} + 1)
    \end{array}
    \]
    It's been proven that a deep Relu network is equivalent to a piece-wise linear function, so the problem of reverse-engineering real-valued neural network is equivalent to find a realization map $\phi$ such that
    \[
    \phi : \mathcal{P}(n_0, \cdots, n_m) \rightarrow PL(\mathbb{R}^{n_0} \rightarrow \mathbb{R})
    \]
    which $PL(\mathbb{R}^{n_0}) \rightarrow \mathbb{R}$ is the set of all real-valued piece-wise linear function defined on $\mathbb{R}^{n_0}$. In order to finding this realization map we limit the $PL(\mathbb{R}^{n_0}) \rightarrow \mathbb{R}$ into set $F_T$ which is the set of bounded piece-wise linear functions.
    \[
    F_T = \{f : f(x) \in PL(R^n \rightarrow R); |f(x)| \leq M ; \forall x\in B_T(0)\}
    \]
    Now we want to find the realization map
    \[
    \phi: \mathbb{R}^D \rightarrow F_T
    \]
    We still need a valid distance to make our space a compact metric space. We consider the following $d_p(.)$ distance:
    \[
    d_p(f, g) = \Big(\int |f(x) - g(x)|^p dx\Big)^{\frac{1}{p}}
    \]
    which satisfies the properties of a valid metric (non-negativity, symmetry and triangle inequality). A special case of this distance is $d_2$ distance which cann be define as follows:
    \[
    d_2(f, g) = \left(\int |f(x) - g(x)|^2 dx\right)^{\frac{1}{2}}
    \]
    We can prove that $(F, d_2)$ is a compact metric space by showing that it is complete and totally bounded which is out of scope of this paper.
    
    \textbf{Def.} Two real-valued networks $\mathcal{N}(x)$ and $\mathcal{N}'(x)$ with same input space $\mathbb{R}^{n_0}$ are isomorphic if they produce the same output for all inputs $x$, i.e., 
    \[
    \mathcal{N}(x) = \mathcal{N}'(x), \forall x \in \mathbb{R}^{n_0}
    \]

    \textbf{Def.} A hyperplane is an $(n-1)$-dimension ($1$-codimension) affine set $H$ as follows:
    \[
    H = \{x \in \mathbb{R}^n | b^T x = \beta\} ; b \in \mathbb{R}^n, \beta \in \mathbb{R}
    \]
    We also need to define the inner product for hyperplanes, we consider the inner product of two hyperplanes $H_1: a_1^Tx=b_1, H_2: a_2^Tx=b_2$ as the intersection between them, as follows:
    \[
    \begin{bmatrix}
    a_1^T \\
    a_2^T 
    \end{bmatrix}
    x = 
    \begin{bmatrix} 
    b_1 \\ 
    b_2
    \end{bmatrix}
    \]
    or we denote it by $Ax = b$, assuming that $||a_1||_2 = ||a_2||_2 = 1$, and $a_1, a_2$ are not colinear, then matrix $A$ is a full row rank matrix and we denote the angle between $a_, a_2$ by $\theta := arccos(a_1^Ta_2)$. if we consider $x_{LN}$ as the least norm solution for equation system $Ax=b$ and $F$ as the reduced row echelon form of $A$ and the permutation matrix $P$ corresponding to this reduced row echelon form, then we have the below set as the intersection between these two hyperplanes. 
    (You can find the math in the appendix $6.1$)
    \[
    <H_1, H_2> := \{
    x_{LN} + P
    \begin{bmatrix}
        -F \\
        I_{n-2}
    \end{bmatrix}
    x'; x'\in \mathbb{R}^{n-2}
    \}
    \]
    
    The bent hyperplane is defined as the set of points where the output of any neuron in the network is zero (the value before non-lienarity $\sigma$ is zero), i.e.,
    \[B_z = \{x \in \mathbb{R}^{n_{0}} | z(x) = 0\}\]
    which $z(x)$ is the pre-activation value for neuron $z$. Let $B$  the union of all bent hyper-planes for all neurons in the network.
    The linear regions of the network correspond to patterns of active and inactive ReLUs, and are defined as the connected components of $\mathbb{R}^{n_{in}}$ that are not in the bent hyperplane.

\section{Method}

    \subsection{Algorithm}
    
    Our reverse-engineering approach begins by sampling points from the input space, so we have $v_1, v_2, \cdots, v_N \in \mathbb{R}^{n_0}$. We know that each point in the domain of a piecewise linear function corresponds to exactly one hyperplane at that point. We can estimate the corresponding hyperplane at an arbitrary point $x$ by calculating the gradient of our ReLU network $f$ as follows, assuming $f$ is differentiable at point $x$:
    \[
    \begin{array}{c}
        \nabla_v f(x) = \frac{d}{dt} f(x + tv) |_{t=0} = \lim_{h\rightarrow 0} \frac{f(x + hv) - f(x)}{h} = \sum_{i=1}^n v_i \frac{\partial f}{\partial x_i} (x); \\
        \nabla f(x) = (\frac{\partial f(x)}{\partial x_1}, \frac{\partial f(x)}{\partial x_1}, \cdots, \frac{\partial f(x)}{\partial x_n})
    \end{array}
    \]
    Similarly, we can have:
    \[
    g(x) = \nabla f(x_i)
    \]
    We assume that $v_1, v_2, \cdots, v_N$ are on the surfaces of some hyperplanes and not in bent areas, i.e., $f$ is differentiable at $v_1, v_2, \cdots, v_N$. So by definition, we know that:
    \[
    g_i (x) = L_i^Tx + b_i
    \]
    for some $L_i \in \mathbb{R}^{n_0}, b_i \in \mathbb{R}$. We use these estimated $g_i$'s to calculate new $g'_i$ with an arbitrary choice of domain and a single constant:
    \[
    g_i'(x) =
    \begin{cases}
    c_i g_i(x) & \text{if } x \in B_{r_i}(v_i) \\
    0 & \text{otherwise}
    \end{cases}
    \]
    We will use these $g'_i$'s to estimate the real-valued ReLU network surface, by using some weights $w_i(x)$ and defining $h$ as follows:
    \[
    h(x) = \sum_{i=1}^{N} w_i(x) g_i'(x)
    \]
    where $\sum_{i=1}^{N} w_i(x) = 1$ for all $i = 1, \ldots, N$. One possible choice for the weighting function is the inverse distance weighting:
    \[
    w(x, x_i) = \frac{1/d(x, v_i)^p}{\sum_{j=1}^{N} 1/d(x, v_j)^p}
    \]
    Define an objective function $L_w(h)$:
    \[
    L_w(h) = \int_{B_T(0)} \lVert h(x) - f(x) \rVert^2 \, dx
    \]
    Despite the interesting properties of the previously introduced weighting function or any other variable-based weighting function, we consider each of them as a single scalar $w_i$. We want to minimize $L_w(h)$ with respect to the weights $w_i$'s. \\
    The second-order partial derivative of $L(h)$ with respect to $w_i$ and $w_j$:
    \[
    \begin{array}{c}
    \frac{\partial^2 L(h)}{\partial w_i \partial w_j} = \frac{\partial}{\partial w_j} \left(\frac{\partial L(h)}{\partial w_i}\right) \\
    \frac{\partial^2 L(h)}{\partial w_i \partial w_j} = \int_{B_T(0)} 2 \left(g'_j(x) g'_i(x) - \frac{\partial f(x)}{\partial w_j} g'_i(x)\right) dx
    \end{array}
    \]
    Since $f(x)$ does not depend on $w_j$, its derivative with respect to $w_j$ is zero, so:
    \[
    \frac{\partial^2 L(h)}{\partial w_i \partial w_j} = \int_{B_T(0)} 2 g'_i(x) g'_j(x) dx
    \]
    The Hessian is as follows:
    \[
    \nabla^2_w L_w(h) =
    \begin{bmatrix}
    \int_{B_T(0)} 2 \cdot (g'_1)^2(x) dx & \int_{B_T(0)} 2 \cdot g'_1(x) \cdot g'_2(x) dx & \dots & \int_{B_T(0)} 2 \cdot g'_1(x) \cdot g'_n(x) dx \\
    \int_{B_T(0)} 2 \cdot g'_2(x) \cdot g'_1(x) dx & \int_{B_T(0)} 2 \cdot (g'_2)^2(x) dx & \dots & \int_{B_T(0)} 2 \cdot g'_2(x) \cdot g'_m(x) dx \\
    \vdots & \vdots & \ddots & \vdots \\
    \int_{B_T(0)} 2 \cdot g'_m(x) \cdot g'_1(x) dx & \int_{B_T(0)} 2 \cdot g'_m(x) \cdot g'_2(x) dx & \dots & \int_{B_T(0)} 2 \cdot (g'_n)^2(x) dx
    \end{bmatrix}
    \]
    We want the optimization of $L_w(h)$ with respect to $w$ to be convex, so the hessian $\nabla^2_w L_w(h)$ should be semi-positive definite. We know that a matrix is SPD if and only if all its eigenvalues is greater or equal than zero.
    We use Gershgorin Circle theorem states which states that every eigenvalue of a square matrix $A$ lies in at least one Gershgorin disk. Each Gershgorin circle $C_i$ is centered at $a_{ii}$ and has a radius equal to the sum of the absolute values of the off-diagonal elements in the $i$-th row:
    \[C_i = \{z \in \mathbb{R} : |z - a_{ii}| \leq R_i\} \ , R_i = \sum_{j \not = i} |a_ij|\]
    In our case we need to have the following conditions to guarantee the convexity of the optimization problem:
    \[
    \int_{B_T(0)} 2 \cdot (g'_i)^2(x) dx \geq  \sum_{j \not = i} \int_{B_T(0)} 2 \cdot g'_i(x) \cdot g'_j(x) dx \ , \ \forall i \\
    \]
    We can prove that the above condition can be reduced to the following condition:
    \[
    (w_i^Tw_i - \sum_{\text{non-zero} \ g_j} w_i^T w_j ) r^2/n  I_n + (b_i^2 - \sum_{\text{non-zero} \ g_j} b_i b_j) r^n V_n \ \ \ \ (\textbf{**})
    \]
    Since the single scalar and the domains of the $g'_i$ functions are arbitrary, we can set the radius of the defining domain of other $g_j$'s such that it satisfies the condition $(\textbf{**})$. We establish different $r$ for each of $g'_i$'s at each row and get the minimum of those $r$ at the end to find the final explicit definitions of $g'_i$'s. This guarantees that the optimization problem is convex.
    
    \subsection{Why do we care about convexity?}
    Convexity is crucial in our method for several reasons. First and foremost, convex functions have a unique global minimum. This means that when optimizing our objective function, we can be confident in having found the best possible solution, rather than getting trapped in a local minimum. This aspect is particularly vital when attempting to reverse engineer a deep ReLU network, as our goal is to discover the most accurate approximation of the target model. Convex optimization problems are generally more robust than their non-convex counterparts. The optimization landscape is smoother, and the algorithms are less sensitive to initialization and hyperparameter choices. Consequently, our method is more likely to perform well in various scenarios and for different types of deep ReLU networks whose structures we aim to uncover. \\
    Additionally, convex optimization problems possess desirable analytical properties that can be exploited to understand the behavior of the optimization algorithm and the solution. For instance, we can derive conditions and constraints that guarantee the convexity of our problem and use them to guide the design of our method. In another study \cite{rolnick2020reverse}, they demonstrated that for two intersecting hyperplane boundaries, $B_z$ and $B_{z'}$:
    \begin{itemize}
        \item $B_z$ bends at their intersection (in which case $z$ occurs in a deeper layer of the network)
        \item $B_{z'}$ bends (in which case $z'$ occurs in a deeper layer)
        \item Neither bends (in which case $z$ and $z'$ occur in the same layer)
    \end{itemize}
    Therefore, by solving the optimization problem, we can gain insights into the structure of the network, particularly by studying the weights and the intersecting domains of $g'_i$'s. This understanding is crucial in reverse engineering the deep ReLU networks and uncovering their hidden structures.
    
    \subsection{One Step Further}
    In our approach to estimating the neural network, we have so far only considered the first-order estimate with respect to $g'_i$'s. However, we can enhance our estimation by including the inner product functions between these hyperplanes, which we previously defined as the intersection area between their corresponding hyperplanes. This inclusion allows us to reformulate the optimization problem as follows:
    \[
    h'(x) = \sum_{i=1}^{N} w_i(x) g_i'(x) + \sum_{i,j} w'_{ij} <g_i'(x), g_j'(x)> 
    \]
    By defining the optimization problem $L_{w, w'}(h')$ and attempting to minimize it with respect to the weights $w$ and $w'$, we can obtain a more accurate estimation of the network's structure. It is important to note, however, that we must establish and satisfy additional constraints to ensure the convexity of this new optimization problem. 
    Incorporating these inner product functions captures higher-order interactions between the hyperplanes, potentially revealing more intricate details about the network's structure. This enriched representation might offer a better approximation of the true function and provide deeper insights into the relationships between the network's components. \\
    In summary, by considering both first-order estimates and inner product functions between the hyperplanes, our method may provide a more comprehensive understanding of the deep ReLU networks' structures. However, to take full advantage of this approach, we must carefully establish and satisfy the necessary constraints to maintain the convexity of the optimization problem, thereby ensuring a unique global minimum and the desirable properties of convex optimization.
    
    \section{Discussion}
        In this work, we propose a novel method for reverse engineering deep ReLU networks by sampling points in the input space, estimating local hyperplanes around these points, and then solving a convex optimization problem to approximate the true function. The central idea behind our approach is to exploit the piecewise linear nature of deep ReLU networks to obtain a reliable approximation of the function represented by the network. \\
        Our main conjecture is that the resulting weights obtained after solving the optimization problem provide valuable information about the hyperplane arrangement of the network and, importantly, the network's architecture. We theorize that by adding the Lasso regularizer $||W||_2$ to the $L_w$ optimization problem, which preserves convexity and smooths the weight solution, the number of non-zero weights will be equal to the number of neurons in the first layer. We aim to prove this conjecture in future work. A more robust hypothesis can be formed by incorporating the inner products of hyperplanes and considering the optimization problem $L_{w, w'}$, assuming we can guarantee its convexity under certain conditions, and adding the Lasso regularizer $||W'||_2$. In this case, the number of non-zero $w'{ij}$ gives us the exact number of activation regions in the space. Previous research only provided an upper bound for this quantity, but our approach, with a properly chosen domain and constant for $g'_i$, can accurately estimate the number of hyperplanes in the space after solving the convex optimization problem that includes the inner product of $g'_i$'s. \\
        Our method involves several steps:
        \begin{itemize}
            \item Sampling points in the input space and querying the black box model to obtain corresponding linear hyperplanes.
            \item Defining a distance metric and a weighting function that help formulate the optimization problem.
            \item Minimizing an objective function that measures the difference between the estimated function and the true function, subject to specific constraints.
        \end{itemize}
        One of the main advantages of our method is its convex nature, which ensures a unique global minimum and simplifies the optimization problem. This convexity allows us to leverage well-established optimization techniques, such as gradient descent, to efficiently find the optimal solution. \\
        Nevertheless, our method has limitations. First, it relies on sampling points in the input space, which may not always provide a comprehensive representation of the true function. This can lead to suboptimal approximations, especially in cases where the input space is high-dimensional or the true function exhibits complex behavior. Moreover, our method assumes that the deep ReLU network has a piecewise linear structure, which may not be valid for other types of activation functions or network architectures. Additionally, the method requires calculating gradients and Hessians, which can become computationally expensive for large-scale problems. \\
        In summary, our proposed method presents a promising framework for reverse engineering deep ReLU networks by leveraging their piecewise linear structure and solving a convex optimization problem. The method's performance depends on the sampling strategy, the choice of distance metric and weighting function, and the optimization technique used. Further research is needed to refine the method and adapt it to other types of networks and activation functions. \\
        We have conducted preliminary practical analyses to test our method on simple ReLU networks, but a detailed discussion of these results is beyond the scope of this paper. In future work, we will examine the main conjecture and the relationship between the structure and optimized weights more closely. We will also attempt to establish convexity guarantees for cases that include inner products. Consequently, to improve the method, future work could explore alternative sampling strategies and incorporate regularization terms to enhance the generalization performance of the approximated function. Furthermore, investigating the relationships between sampled hyperplanes and integrating these relationships into the optimization problem, as in the case of inner product terms, may lead to improved performance in approximating the



\small

\newpage

\nocite{*} 
\bibliographystyle{achemso}
\bibliography{ref} 


\newpage

\section{Supplementary Material}

\subsection{Inner Product}
Suppose we have two $(n-1)$-dimension hyperplanes in $\mathbb{R}^n$
\[
    \begin{array}{c}
        H_1: a_{11}x_1 + a_{12}x_2 + \cdots + a_{1n}x_n = a_1^Tx = b_1 \\
        H_2: a_{21}x_1 + a_{22}x_2 + \cdots + a_{2n}x_n = a_2^Tx = b_2 \\
    \end{array}
\]
We write them in matrix form as follows:
\[
    \begin{bmatrix}
        a_1^T \\
        a_2^T 
    \end{bmatrix}
    x = 
    \begin{bmatrix} 
        b_1 \\ 
        b_2
    \end{bmatrix}
\]
or we denote it by $Ax = b$, assuming that $||a_1||_2 = ||a_2||_2 = 1$, and $a_1, a_2$ are not colinear, then matrix $A$ is a full row rank matrix and we denote the angle between $a_, a_2$ by $\theta := arccos(a_1^Ta_2)$. \\
The solution set of the linear system $Ax=b$ is a $(n-2)$-dimensional affine space.
To find this affine space, we must find a particular solution and the null space of $A$. One particular solution would be the least-norm solution, i.e., the one closest to the origin:
\[
\begin{array}{c}  
x_{\text{LN}} :=  A^T \left(  A  A^\top \right)^{-1}  b = \\
\begin{bmatrix}
    | & | \\  
    a_1 &  a_2\\ 
    | & |
    \end{bmatrix}
\begin{bmatrix} \|  a_1 \|_2^2 & \langle  a_1,  a_2 \rangle\\ \langle  a_2,  a_1 \rangle & \|  a_2 \|_2^2
\end{bmatrix}^{-1} 
\begin{bmatrix}
    b_1\\
    b_2
\end{bmatrix} = 
\begin{bmatrix}
    a_1 &  a_2
\end{bmatrix} 
\begin{bmatrix} 
    1 & \cos (\theta)\\
    \cos (\theta) & 1
\end{bmatrix}^{-1} 
\begin{bmatrix} 
    b_1\\ 
    b_2
\end{bmatrix} \\
=\frac{1}{\sin^2 (\theta)} 
\begin{bmatrix} 
    a_1 & a_2
\end{bmatrix} 
\begin{bmatrix} 
    1 & -\cos (\theta)\\
    -\cos (\theta) & 1
\end{bmatrix} 
\begin{bmatrix} 
    b_1\\ 
    b_2
\end{bmatrix}\\
\Rightarrow x_{LN}
= \left(\frac{b_1 - b_2 \cos (\theta)}{\sin^2 (\theta)}\right) a_1 + \left(\frac{b_2 - b_1 \cos (\theta)}{\sin^2 (\theta)}\right) a_2
\end{array}
\]
To find the ($n-2$)-dimensional null space of $A$, we solve the homogeneous linear system $Ax=0_2$. Introducing an $n\times n$ permutation matrix $P$ with certain desired properties, we reorder the columns of $A$, i.e., $APP^Tx=0_2$, and obtain a homogeneous linear system in $y:=P^Tx$
\[
APy = 0_2
\]
We chose $P$ such that it generated the reduced row echelon form with $2 \times 2$ matrix $E$, as follows:
\[
EAP = 
\begin{bmatrix}
    I_2 & F
\end{bmatrix}
\]
So the null space of $AP$ is given by:
\[
\{\tau \in \mathbb{R}^{n-2} |
\begin{bmatrix}
    -F \\
    I_{n-2}
\end{bmatrix} \tau
\}
\]
and the null space of $A$ is:
\[
\{\tau \in \mathbb{R}^{n-2} |
P \begin{bmatrix}
    -F \\
    I_{n-2}
\end{bmatrix} \tau
\}
\]
Hence the intersection and the desired inner product is like below:
\[
<H_1, H_2> := \{
\tau \in \mathbb{R}^{n-2} |
x_{LN} + P
\begin{bmatrix}
    -F \\
    I_{n-2}
\end{bmatrix}
\tau
\}
\]

\subsection{Hessian Calculations}
Let's compute the partial derivative of $L(h)$ with respect to $w_i$:
\[
\frac{\partial L(h)}{\partial w_i} = \frac{\partial}{\partial w_i} \left(\int_{B_T(0)} ||h(x) - f(x)||^2 dx\right)
\]
Using the chain rule:
\[
\frac{\partial L(h)}{\partial w_i} = \int_{B_T(0)} 2(h(x) - f(x)) \frac{\partial(h(x) - f(x))}{\partial w_i} dx
\]
Since $f(x)$ does not depend on $w_i$, its derivative with respect to $w_i$ is zero:
\[
\frac{\partial(h(x) - f(x))}{\partial w_i} = \frac{\partial h(x)}{\partial w_i}\]
Now, we need to compute the derivative of $h(x)$ with respect to $w_i$:
\[\frac{\partial h(x)}{\partial w_i} = \frac{\partial}{\partial w_i} \left(\sum_j w_j g'_j(x)\right) = g'_i(x)
\]
Finally, substituting this back into the partial derivative of $L(h)$ with respect to $w_i$:
\[
\frac{\partial L(h)}{\partial w_i} = \int_{B_T(0)} 2(h(x) - f(x)) g'_i(x) dx
\]
Now let's compute the second-order partial derivative of $L(h)$ with respect to $w_i$ and $w_j$:
\[
\frac{\partial^2 L(h)}{\partial w_i \partial w_j} = \frac{\partial}{\partial w_j} \left(\frac{\partial L(h)}{\partial w_i}\right)
\]
Using the chain rule:
\[
\frac{\partial^2 L(h)}{\partial w_i \partial w_j} = \int_{B_T(0)} 2 \left(g'_j(x) g'_i(x) - \frac{\partial f(x)}{\partial w_j} g'_i(x)\right) dx
\]
Since $f(x)$ does not depend on $w_j$, its derivative with respect to $w_j$ is zero:
\[
\frac{\partial^2 L(h)}{\partial w_i \partial w_j} = \int_{B_T(0)} 2 g'_i(x) g'_j(x) dx
\]
\[
H_{i,j}(x) = \frac{\partial^2 L(h)}{\partial w_i \partial w_j} = \int_{B_T(0)} 2 \cdot g'_i(x) \cdot g'_j(x) dx
\]
In matrix form, the Hessian matrix can be written as:
\[\begin{bmatrix}
\frac{\partial^2 L(h)}{\partial w_1^2} & \frac{\partial^2 L(h)}{\partial w_1 \partial w_2} & \dots & \frac{\partial^2 L(h)}{\partial w_1 \partial w_m} \\
\frac{\partial^2 L(h)}{\partial w_2 \partial w_1} & \frac{\partial^2 L(h)}{\partial w_2^2} & \dots & \frac{\partial^2 L(h)}{\partial w_2 \partial w_m} \\
\vdots & \vdots & \ddots & \vdots \\
\frac{\partial^2 L(h)}{\partial w_m \partial w_1} & \frac{\partial^2 L(h)}{\partial w_m \partial w_2} & \dots & \frac{\partial^2 L(h)}{\partial w_m^2}
\end{bmatrix} = \]
\[
\begin{bmatrix}
\int_{B_T(0)} 2 \cdot (g'_1)^2(x) dx & \int_{B_T(0)} 2 \cdot g'_1(x) \cdot g'_2(x) dx & \dots & \int_{B_T(0)} 2 \cdot g'_1(x) \cdot g'_n(x) dx \\
\int_{B_T(0)} 2 \cdot g'_2(x) \cdot g'_1(x) dx & \int_{B_T(0)} 2 \cdot (g'_2)^2(x) dx & \dots & \int_{B_T(0)} 2 \cdot g'_2(x) \cdot g'_m(x) dx \\
\vdots & \vdots & \ddots & \vdots \\
\int_{B_T(0)} 2 \cdot g'_m(x) \cdot g'_1(x) dx & \int_{B_T(0)} 2 \cdot g'_m(x) \cdot g'_2(x) dx & \dots & \int_{B_T(0)} 2 \cdot (g'_n)^2(x) dx
\end{bmatrix}\]

\subsection{Convexity guarantee}
As we discussed using the Gershgorin circle theorem we need to have the following conditions to guarantee the convexity of the optimization problem:
\[
\int_{B_T(0)} 2 \cdot (g'_i)^2(x) dx \geq  \sum_{j \not = i} \int_{B_T(0)} 2 \cdot g'_i(x) \cdot g'_j(x) dx \ , \ \forall i \\
\]
Now at each point $x$ only a few of $g_i$'s are non-zero we can write the integral of two linear functions over a ball like below:
if $g_i(x) = w_i^T (x - x_i) + b_i$ and $g_j(x) = w_j^T (x-x_j) + b_j$ Then, the integral of their product over an $n$-dimensional ball with radius $r$ can be computed as follows:
\[
\int_{B_r} g_i(x) g_j(x) dx 
= \int_{B_r} (w_i^T x + b_i)(w_j^T x + b_j) dx \]
\[
= \int_{B_r} (w_i^T x)(w_j^T x) + b_i(w_j^T x) + b_j(w_i^T x) + b_i b_j dx \]
\[= \int_{B_r} (w_i^T x)(w_j^T x) dx + b_i \int_{B_r} (w_j^T x) dx + b_j \int_{B_r} (w_i^T x) dx + b_i b_j \int_{B_r} 1 dx \]
Since $w_i$ and $w_j$ are constant vectors, we can take them outside the integral:
\[= w_i^T w_j \int_{B_r} x x^T dx + b_i (w_j^T \int_{B_r} x dx) + b_j (w_i^T \int_{B_r} x dx) + b_i b_j \int_{B_r} 1 dx\]
Next, we can compute the integrals separately. Using the formula for the volume of an n-dimensional ball, we can write:
\[\int_{B_r} x x^T dx = r^2/n  I_n\]
where $I_n$ is the n-dimensional identity matrix. Using this, we can simplify the integral of the product of the linear functions as:
\[\int_{B_r} g_i(x) g_j(x) dx = w_i^T w_j  r^2/n  I_n + b_i (w_j^T r^2/n  0) + b_j (w_i^T r^2/n  0) + b_i b_j  r^n  V_n\]
where $V_n$ is the volume of the $n$-dimensional ball with radius r centered at the origin. Since the integrals of the linear functions with respect to x are zero due to symmetry, the second and third terms above are zero. The fourth term is simply the volume of the ball times the product of the biases, and the first term is a scaled version of the n-dimensional identity matrix times the product of the weight vectors. \\
Therefore, the integral of the product of two linear functions $g_i(x)$ and $g_j(x)$ over an n-dimensional ball with radius r is given by:
\[\int_{B_r} g_i(x) g_j(x) dx = w_i^T w_j  (r^2/n)  I_n + b_i b_j r^n V_n\]

Now going back to the hessian , we have:

\[
(w_i^Tw_i - \sum_{\text{non-zero} \ g_j} w_i^T w_j ) r^2/n  I_n + (b_i^2 - \sum_{\text{non-zero} \ g_j} b_i b_j) r^n V_n \ \ \ \ (\textbf{**})
\]

So we prove that we can establish a bound for making the optimization problem convex.

\end{document}